\documentclass[letterpaper, 10 pt, conference]{ieeeconf}  

\IEEEoverridecommandlockouts                              

\usepackage{graphics} 
\usepackage{amsmath} 
\usepackage{amssymb}  

\title{\LARGE \bf
Gaussian Processes for Modelling Spatial Fields with Robot Swarms
}

\author{Guillermo Legarda Herranz$^{1}$, Gianpiero Francesca$^{2}$ and Mauro Birattari$^{1}$
\thanks{$^{1}$Guillermo Legarda Herranz and Mauro Birattari are with IRIDIA,
        Université libre de Bruxelles, 1050 Brussels, Belgium
        {\tt\small guillermo.legarda.herranz@ulb.be}, {\tt\small mauro.birattari@ulb.be}}%
\thanks{$^{2}$Gianpiero Francesca is with Toyota Motor Europe
        {\tt\small gianpiero.francesca@toyota-europe.com}}%
}

\begin{document}

\maketitle
\thispagestyle{empty}
\pagestyle{empty}

\begin{abstract}

Robot swarms, by virtue of their decentralised architecture, are a natural tool for scalable, robust modelling of spatial fields, such as water temperature, wind velocity, or terrain elevation.
However, existing methods rely on external positioning systems that allow each robot to determine its own position in space.
Here, we introduce location-unaware Gaussian process regression (LU-GPR) as a solution to the modelling of spatial fields in the absence of such positioning systems.
LU-GPR allows each robot to infer the posterior mean and variance of the field in space, while simultaneously agreeing on a common frame of reference with its peers, using only local sensing and communication.
We propose an online algorithm that allows each robot to consistently infer local estimates as its local frame of reference converges to the common one.
By means of a product of experts model, each robot also combines the estimates of its peers with its own to obtain a global model.
Our results show that LU-GPR scales well with the number of robots and is robust to limited communication ranges.
We also demonstrate how it can be used in real-world monitoring scenarios to estimate the flow of an evacuating crowd.

\end{abstract}

\section{INTRODUCTION}
Spatial fields arise in a large variety of natural phenomena.
We may think of ocean currents and temperature variations, wind patterns, or terrain elevations, as some of the traditionally studied examples.
Spatial fields also arise in the form of collective motion patterns, such as in human crowds or traffic flow.
The ability to model them is, therefore, of great interest in the fields of oceanography~\cite{MolBreSuk2015icra}, topography~\cite{BelLabSkr2012robio}, and urban planning~\cite{PanLiLiu-etal2026ARCRAS}, and an important challenge for robotic systems.

Gaussian process regression (GPR) provides an effective framework for modelling spatial fields~\cite{WilRas20006book}.
Following a Bayesian approach, GPR infers a distribution over the underlying process, which can be recursively exploited for efficient sampling and navigation~\cite{SchSpeKra2018JMP}.
Unfortunately, in the standard formulation of GPR, inference is an $\mathcal{O}(n^3)$ operation in the number of collected samples, which makes it impractical for large-scale applications (see Section \ref{sec:gpr}).

Scalable Gaussian processes (GPs) have emerged as a solution to this problem~\cite{LiuOngSheCai2020TNNLS}.
The main idea behind monolithic, scalable GPs is to build sparse approximations of either the data or the GP itself, as a way to bound the computational complexity of the inference process~\cite{Tit2009aistats,LazQuiRasFig2010JMLR,HayImaYos2020aistats}.
Alternatively, learning and inference can be performed in a distributed manner~\cite{DeiNg2015icml}.
In this scenario, a set of independent experts process a subset of the collected data, which is then used by a centralised computer to infer a global model.

The development of distributed approaches to GPR opened the door for multi-robot implementations.
In multi-robot systems, a fully decentralised learning and inference architecture can be obtained by omitting the centralised computer.
Hoang et al.~\cite{HoaHoaLowHow2019aaai} demonstrated how decentralised architectures can lead to efficient learning and scale to many learning agents.
Norton et al.~\cite{NorStaWarPet2023JIRS} introduced an approach based on clustering input data to estimate random vector fields.
Llorente et al.~\cite{LloWaxDju2025icassp} (and~\cite{LloWasJan-etal2025arxiv}) introduced a robust, adaptive approach to online learning based on an approximation of the kernel function of the GP.
Pratissoli et al.~\cite{PraManProSab2025TRO} developed an adaptive method that can continuously learn spatial fields that change over time, and went as far as demonstrating its behaviour on a system of real TurtleBot3 Burger robots.

It is therefore clear that multi-robot systems can successfully integrate GPs for scalable, decentralised regression.
However, research on decentralised GPR has been limited to settings where the learning agents have access to a positioning system.
In swarm robotics, we do not make this assumption.
To ensure the scalability of the swarm, robots rely only on local perception of the environment, and do not have access to global information, such as position data~\cite{Sah2005sab}.
This allows robot swarms to operate under GPS-denied conditions without compromising their performance.

In this work, we present location-unaware GPR (LU-GPR), an approach to fully decentralised, online GPR that does not rely on any external positioning system.
We exploit recent advances in Gaussian belief propagation (GBP), which allow the robots in a swarm to converge towards a common frame of reference for position estimation through local communication and iterative message passing~\cite{JonHau2025AR}.
As the individual robots have no means to determine when the GBP algorithm has converged, LU-GPR allows for simultaneous position estimation and GP regression.
LU-GPR is based on an incremental algorithm that uses a sparse approximation of the kernel function in a GP using random Fourier features~\cite{GijMet2013NN}, allowing inference complexity to depend on the number of random features rather than on the number of collected samples.
It is also designed to be robust to unreliable samples and adaptive, allowing it to forget older samples in favour of more recent ones. 

To demonstrate the suitability of LU-GPR to estimate real-world spatial fields, we consider an evacuation scenario, where crowds are prone to exhibit collective motion patterns~\cite{CorTos2023ARCMP}.
In an indoor environment, a swarm of ground robots could sample pedestrian velocities to estimate the global crowd behaviour.
These estimates could then be communicated to external security personnel, or used by the swarm itself to detect and prevent bottlenecks, oscillations, and other, potentially dangerous emergent phenomena~\cite{PanLiLiu-etal2026ARCRAS}.

\section{MATHEMATICAL BACKGROUND}
In this section, we introduce the mathematical concepts needed to understand our contribution.
A complete presentation of these ideas is beyond the scope of this work, and we refer the interested reader to references in the corresponding sections for more comprehensive reviews.

\subsection{Gaussian process regression}\label{sec:gpr}
Gaussian process regression (GPR) is a non-parametric Bayesian approach to estimating a latent function from observation data~\cite{WilRas20006book,SchSpeKra2018JMP}.
Consider a set $\mathcal{D}$ of observations, $\mathcal{D} = \{(\mathbf{x}_i, y_i)\}$, where $\mathbf{x}_i \in \mathbb{R}^D$ denotes the input vector indexed by the value of $i \in \{1, 2, \dots, n\}$, and $y_i$ represents the observed output, here assumed to be scalar.
In GPR, we assume a linear model,
\begin{align}\label{eq:linearmodel}
    y_i = f(\mathbf{x}_i) + \epsilon, &\qquad f(\mathbf{x}_i) = \phi(\mathbf{x}_i) \mathbf{w},
\end{align}    
where $\epsilon \sim \mathcal{N}(0,\sigma_{n}^2)$ captures the noise of the observations.
The function $\phi(\mathbf{x}_i)$ maps each input vector into an $m$-dimensional feature space, and the vector $\mathbf{w}$ represents the weights that we aim to learn from the data.

Given $\mathcal{D}$, we define $X = [\mathbf{x}_1, \dots, \mathbf{x}_n]^\text{T}$ as the matrix of all observed inputs, and $\mathbf{y} = [y_1, \dots, y_n]^\text{T}$ as the vector of all observed outputs, such that $\mathcal{D} = \{X, \mathbf{y}\}$.
In the weight-space view of GPR, we place a prior distribution on the weights, $\mathbf{w} \sim \mathcal{N}(\mathbf{0},\Sigma_p)$, where $\Sigma_p$ is a positive-definite covariance matrix. 
Using Bayes' theorem and integrating over all possible $\mathbf{w}$, we obtain the predictive distribution
\begin{align}\label{eq:weightspace}
    p(f_*|\mathbf{x}_*,\mathcal{D}) = \mathcal{N}(&\phi^\text{T}(\mathbf{x}_*) A^{-1} \Phi^\text{T} \mathbf{y}, \nonumber \\
    &\sigma_{n}^2 \phi^\text{T}(\mathbf{x}_*) A^{-1} \phi(\mathbf{x}_*),
\end{align}
where $f_*$ represents the predicted value of the latent function at the test input vector, $\mathbf{x}_*$, $A = \Phi^\text{T} \Phi + \sigma_{n}^2 \Sigma_p^{-1}$, and $\Phi = \phi(X) = [\phi(\mathbf{x}_1), \dots, \phi(\mathbf{x}_n)]^\text{T}$.

When the feature-space mapping is unknown, or its dimensionality is too large (potentially infinite), we can replace all occurrences of its inner products by evaluations of a kernel function, $k(\mathbf{x}, \mathbf{x}')$.
This kernel trick leads to the function-space view of GPR~\cite{WilRas20006book}.
Both in the weight-space and function-space formulations of GPR, computing the predictive posterior involves calculating the inverse of a matrix whose size grows with the size of the data set.
This is the most significant bottleneck in GPR, as matrix inversion is an $\mathcal{O}(n^3)$ operation.

Stationary kernels, where $k(\mathbf{x}, \mathbf{x}') = k(\mathbf{x}-\mathbf{x}')$, can be approximated using a finite number of random Fourier features~\cite{LazQuiRasFig2010JMLR, RahRec2007nips}.
Specifically, we can define $k(\mathbf{x},\mathbf{x}') \approx \phi^\text{T}(\mathbf{x}) \phi(\mathbf{x}')$, where
\begin{equation}\label{eq:features}
    \phi(\mathbf{x}) = \frac{\sigma_f}{\sqrt{m}} [\mathbf{z}^\text{T}_{\omega_1}(\mathbf{x}), \dots, \mathbf{z}^\text{T}_{\omega_m}(\mathbf{x})]^\text{T}.
\end{equation}
Here, $\phi(\mathbf{x}) \in \mathbb{R}^{2m}$, $\sigma_f^2$ is the signal variance, and
\begin{equation}\label{eq:sample}
    \mathbf{z}_{\omega_j}(\mathbf{x}) = [\cos(\omega_j^T \mathbf{x}), \sin(\omega_j^T \mathbf{x})]^\text{T}.
\end{equation}
Each frequency, $\omega_j$, is sampled from the spectral density of the kernel.
We can then use \eqref{eq:weightspace} for inference, where matrix inversion is now $\mathcal{O}(m^3)$.
The resulting sparse approximation does not scale with the data, and is therefore suitable for big-data applications.

The incremental sparse spectrum GPR algorithm (I-SSGPR) provides update rules for the matrix $A$ and the vector $\mathbf{b} := \Phi^\text{T} \mathbf{y}$ in \eqref{eq:weightspace} as data samples are gathered in time, allowing further reduction of the computational complexity of inference to $\mathcal{O}(m^2)$~\cite{GijMet2013NN}.
Consider now that the data in \eqref{eq:linearmodel} is indexed by time, that is, $i=t$.
When a new sample arrives, $(\mathbf{x}_t, y_t)$, we define $\Phi_t = [\Phi^T_{t-1}, \phi_t]^\text{T}$, where $\phi_t = \phi(\mathbf{x}_t)$, and $\mathbf{y}_t = [\mathbf{y}^\text{T}_{t-1}, y_t]^\text{T}$.
The values of $A$ and $\mathbf{b}$ are then updated as
\begin{align}
    A_t &= A_{t-1} + \phi_t \phi_t^\text{T}, \label{eq:aupdate}\\
    \mathbf{b}_t &= \mathbf{b}_{t-1} + \phi_t y_t, \label{eq:bupdate}
\end{align}
where $A_0 = \sigma_n^2 \Sigma_p^{-1}$ and $\mathbf{b}_0 = \mathbf{0}$.
Instead of keeping track of the full matrix $A_t$, I-SSGPR stores the value of its upper triangular Cholesky factor, $U_t$, such that $A_t = U_t^\text{T} U_t$.
The value of $U_t$ is updated by means of a thin QR factorisation of the matrix $\tilde{U}_t = [U_t^\text{T}, \phi_t]^\text{T}$~\cite{GolVan2013book}.
The predictive mean and variance are then computed using back and forward substitutions.

\subsection{Gaussian belief propagation}
Gaussian belief propagation (GBP) is a method for performing marginal inference on a factor graph using message passing~\cite{DavOrt2019arxiv, OrtEvaDavi2021arxiv}.
Factor graphs are bipartite graphs whose nodes are either variables or factors.
Variables represent quantities of interest that we want to estimate, but which are not directly observable.
Factors connect variables and constrain their values based either on observations of the environment, or on specified priors.

In the standard GBP algorithm, the marginal posterior of each variable can be calculated through iterative message passing.
At each iteration, the algorithm updates the belief at a variable node, given as a Gaussian distribution, using all its neighbouring factors.
In tree-structured graphs, this will eventually converge to the exact marginals, but GBP gives empirically good results even in the presence of cycles.

Jones and Hauert~\cite{JonHau2025AR} recently implemented GBP in robot swarms.
In their work, robots operate without global positioning and with limited-range communication only, and GBP allows them to converge to a common frame of reference.
In their implementation, each robot keeps a bounded factor graph of variables representing its position with respect to the common frame, where positions are calculated from odometry measurements. 
As robots meet their neighbours, they create outward-facing factors that constrain their mutual factor graphs, allowing local position estimates to converge to a consistent representation.

\section{SPATIAL FIELD MODELLING}
We now introduce LU-GPR, our proposed approach to modelling spatial fields with a robot swarm, where the robots have no knowledge of their position in space, and can only communicate within a limited range.
Each robot runs the GBP algorithm of Jones et al.~\cite{JonHau2025AR}, while simultaneously updating its GP model using data collected in time, as we describe in Section \ref{sec:luissgpr}.
To build a global model of the latent function, robots collect the models of their peers and fuse them with their own, as we describe in Section \ref{sec:gpoe}.

\subsection{Location-unaware GPR}\label{sec:luissgpr}
Consider a robot that runs the GBP algorithm and samples the spatial field at fixed time intervals.
As its local frame of reference converges to one shared with its peers, the positions at which all previous samples were collected change accordingly.
Additionally, the number of samples might be arbitrarily large, but the robot only keeps a bounded number of variable nodes in its factor graph.
Therefore, the robot cannot rely on the GBP algorithm to keep an accurate estimate of the locations where the samples where collected, whose uncertainty will grow  as the robot moves.

This scenario motivates the need for an online GPR algorithm that can adjust the values of functions that depend on stored inputs, $X$, as well as assign a lower weight to older samples.
Algorithms that fit the latter criterion are known as adaptive algorithms~\cite{VanLazSan2012TNNLS,PraManProSab2025TRO,LloWasJan-etal2025arxiv}.
Additionally, in real-world scenarios, noisy sensing can lead to unreliable measurements, so we are also interested in robust solutions.

To introduce robustness and adaptivity into our model, we propose a weighted linear model with a time-dependent weighting function~\cite{SheXiaChe-etal2018TIE},
\begin{align}\label{eq:model}
    y_i = f(\mathbf{x}_i) + \frac{\epsilon}{g_i(t)}, &\qquad g_i(t) = \lambda^{(t - t_i)}\rho_i,
\end{align}
where $\lambda \in (0, 1]$ is a forgetting factor, $t_i$ is the time at which the sample $(\mathbf{x}_i, y_i)$ is collected, and $\rho_i > 0$ is the sample weight.
The effect of the $\lambda$-factor in $g_i(t)$ is, therefore, to make earlier samples less reliable by scaling up the variance of the noise, $\epsilon$.
All other variables are defined as in \eqref{eq:linearmodel}.

The weighted linear model leads to the following predictive posterior,
\begin{align}\label{eq:pred}
    p(f_*|\mathbf{x}_*,t_*,\mathcal{D}) = \mathcal{N}(&\phi^\text{T}(\mathbf{x}_*) B^{-1}(t_*) \Phi^\text{T} G^2(t_*) \mathbf{y}, \nonumber \\
    & \sigma_n^2 \phi^\text{T}(\mathbf{x}_*) B^{-1}(t_*) \phi(\mathbf{x}_*)),
\end{align}
where $B(t_*) = \Phi^\text{T} G^2(t_*) \Phi + \sigma_n^2 \Sigma_p^{-1}$, and $G(t_*) = \text{diag}([g_1(t_*), \dots, g_n(t_*)])$.
The derivation is identical to that by Shen et al.~\cite{SheXiaChe-etal2018TIE}, except for the conditioning on the test time, $t_*$.
In the following, we will omit the dependence of $B$ and $G$ on $t_*$, for convenience.

We now seek incremental update rules similar to \eqref{eq:aupdate} and \eqref{eq:bupdate} that allow us to solve \eqref{eq:pred} recursively as new samples arrive, without having to store every sample.
These update rules must also correct the value of $\Phi$, as it depends on all previous inputs, $X$.

To address the correction of $\Phi$, we consider the effect of a finite displacement of the input vector, $\tilde{\mathbf{x}} = \mathbf{x} + \mathbf{\delta x}$.
As we are concerned with position data, we restrict our formulation to the two-dimensional case, $\mathbf{x} \in \mathbb{R}^2$.
Substituting $\tilde{\mathbf{x}}$ into \eqref{eq:sample} and using the trigonometric identities for sines and cosines of sums of angles, we obtain
\begin{align}
    \mathbf{z}_{\omega_j}(\tilde{\mathbf{x}}) &= [\cos(\omega_j^\text{T} \mathbf{x}) \cos(\omega_j^\text{T} \mathbf{\delta x}) - \sin(\omega_j^\text{T} \mathbf{x}) \sin(\omega_j^\text{T} \mathbf{\delta x}), \nonumber \\
    &\qquad \sin(\omega_j^\text{T} \mathbf{x}) \cos(\omega_j^\text{T} \mathbf{\delta x}) + \cos(\omega_j^\text{T} \mathbf{x}) \sin(\omega_j^\text{T} \mathbf{\delta x})]^\text{T} \nonumber \\
    &= R_j(\omega_j^\text{T} \mathbf{\delta x}) \mathbf{z}_{\omega_j}(\mathbf{x}),
\end{align}
where $R_j(\cdot)$ is the two-dimensional rotation matrix.
Clearly, a finite displacement in space corresponds to a finite phase shift, and can therefore be represented by a rotation of the unit vector $\mathbf{z}_{\omega_j}(\mathbf{x})$.

\begin{figure}
    \centering
    \includegraphics{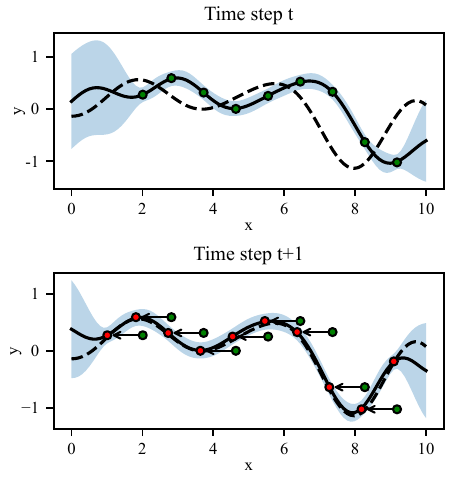}
    \caption{Illustration of the online data correction procedure. Dashed lines represent the function to be estimated; solid lines, the posterior mean; and shaded regions bound the values within one standard deviation of the posterior mean. At time step $t$, the local frame of reference is at $-1$, and therefore all sampled data appear shifted 1 unit to the right in the global frame of reference. At time step $t+1$, the local frame of reference is aligned with the global one, and all previously collected samples are shifted accordingly to accommodate the new sample collected at $x=9$.}
    \label{fig:correction}
\end{figure}

We can define the block diagonal matrix $R = \text{blockdiag}(\{R_j, \dots, R_m\})$, where we use $R_j = R_j (\omega_j^\text{T} \mathbf{\delta x})$.
This allows us to update $\Phi$ as $\Phi \leftarrow \Phi R^\text{T}$.
As each block in $R$ is of fixed, $2 \times 2$ size, this update can be computed in $\mathcal{O}(m)$ time.
At time $t$, we combine the arrival of a new sample with the displacement of the origin since the last update, and define
\begin{equation}\label{eq:phiupdate}
    \Phi_t = [R \Phi^\text{T}_{t-1}, \phi_t]^\text{T},
\end{equation}
where $\phi_t = \phi(\mathbf{x}_t)$.
We illustrate how this correction affects consecutive estimates in Fig. \ref{fig:correction}.

For the update of the weight matrix, $G$, we first notice from \eqref{eq:model} that $g_i(t) = \lambda g_i(t-1)$.
Furthermore, when a sample arrives at time $t$, then $t_i = t$, and $g_i(t) = \rho_i$.
We can then define
\begin{equation}\label{eq:gupdate}
    G_t = \begin{bmatrix}
        \lambda G_{t-1} & \mathbf{0} \\
        \mathbf{0}^\text{T} & \rho_i
    \end{bmatrix}.
\end{equation}

In the following, we assume that the weights are independent, identically distributed, $\Sigma_p = \sigma_p^2 I$, with variance $\sigma_p^2$.
Using \eqref{eq:phiupdate} and \eqref{eq:gupdate}, we can then define the update rules for $B$ and $\mathbf{c} := \Phi^\text{T} G^2 \mathbf{y}$ as
\begin{align}
    B_t &= \lambda^2 R B_{t-1} R^\text{T} + \rho_i^2 \phi_t \phi_t^\text{T} + Q \label{eq:varupdate} \\
    \mathbf{c}_t &= \lambda^2 R \mathbf{c}_{t-1} + \rho_i^2 \phi_t y_t, \label{eq:meanupdate}
\end{align}
where $B_0 = \sigma_n^2 \Sigma_p^{-1}$, $\mathbf{c}_0 = \mathbf{0}$, and $Q = (1 - \lambda^2) \sigma_n^2 \Sigma_p^{-1}$ is a symmetric, positive-definite constant matrix that can be pre-computed (see Appendix).

These rules have some interesting properties.
First, in the absence of a new sample ($\phi_t = \mathbf{0}$), or any displacement of the local frame of reference ($R = I$), we have pure forgetting, with $B_t = \lambda^2 B_{t-1} + (1 - \lambda^2) \sigma_n^2 \Sigma_p^{-1}$, and $\mathbf{c}_t = \lambda^2 \mathbf{c}_{t-1}$.
With the exception of the additional $\lambda$ factor in the update of $\mathbf{c}_t$, these expressions recover the back-to-prior (B2P) forgetting, such that, as $\lambda \rightarrow 0$, we lose all information acquired by the GP, and the posterior in \eqref{eq:pred} approaches the prior~\cite{VanLazSan2012TNNLS}.
On the other hand, if all we have is a displacement of the local frame of reference (now $\lambda = 1$), then $B_t = R B_{t-1} R^\text{T}$.
As $R^\text{T}=R^{-1}$, the update of $B_t$ is a similarity transformation, and $B_t$ and $B_{t-1}$ are similar~\cite{GolVan2013book}.
At the same time, $\mathbf{c}_t = R \mathbf{c}_{t-1}$, so the mean is rotated in pairs of two elements.

In a similar fashion to Gijsberts and Metta~\cite{GijMet2013NN}, we can define update rules for $U_t$, the upper-triangular Cholesky factor of $B_t$.
From \eqref{eq:varupdate},
\begin{align}
    B_t = U_t^\text{T} U_t &= (\lambda R U^\text{T}_{t-1}) (U_{t-1} R^\text{T} \lambda) + (\rho_i \phi_t) (\phi_t^\text{T} \rho_i) \nonumber \\
    & \quad +\ Q^{1/2} (Q^{1/2})^\text{T}\nonumber \\
    &= \tilde{U}_{t-1}^\text{T} \tilde{U}_{t-1},
\end{align}
where $\tilde{U}_{t-1} = [\lambda R U^\text{T}_{t-1}, \rho_i \phi_t, Q^{1/2}]^\text{T}$.
The updated Cholesky factor can then be calculated using a thin QR factorisation, or a rank-$(2m+1)$ Cholesky update.
As $Q$ is a full-rank matrix with the same dimensions as $B$, both of these options, as well as \eqref{eq:varupdate}, can be computed in $\mathcal{O}(m^3)$, which does not scale with the number of data samples~\cite{GolVan2013book}.
Therefore, while this Cholesky-based update does not reduce the computational complexity of the update, it does prevent the instability problems of direct matrix inversion.

\subsection{Ensemble learning}\label{sec:gpoe}
At regular intervals, each robot randomly selects one of its neighbours, which are the robots within its communication range, and obtains a local model from it, where a model consists of the calculated posterior mean and variance.
We assume that each robot has a unique identifier, and that it keeps a time stamp on its local model.
This time stamp does not need to be synchronised with the rest of the swarm; only incremented every time the model is updated.
To maximise the information gained from each interaction, the robot first sends to its selected neighbour a list of the identifiers of all the robots whose model it has received, and their corresponding time stamps.
The responding robot first verifies if it has any models that the other robot does not have, in which case it selects one of those models at random, and sends back a message containing the model, its time stamp, and the identifier of the robot that calculated it.
If the responding robot does not have any new models, it sends back the one with the largest time stamp difference.

This communication protocol operates at the same rate as the messaging required by the GBP algorithm, but we keep both implementations independent for modularity.
With every regression update, each robot uses \eqref{eq:varupdate} and \eqref{eq:meanupdate} to update its own model.
However, it must also keep all received models aligned with its changing, local frame of reference, and apply the appropriate forgetting due to odometry drift.
Therefore, each robot updates each received model using \eqref{eq:varupdate} and \eqref{eq:meanupdate} with $\phi_t = \mathbf{0}$.
Doing so would require that each robot also shares the sampled frequencies to compute $R$ accordingly.
Instead, we assume that each robot is capable of sampling the same frequencies through a common seed, and use the same $R$ for all models.

As the robots operate with a limited communication range, each of them only has access to partial set of observations.
We define the complete, system-wide data set, as $\mathcal{D} = \{\mathcal{D}^{(1)}, \dots, \mathcal{D}^{(k)}\}$, where $k$ is the number of robots in the swarm, and $\mathcal{D}^{(k)}$ is the set of observations collected by the $k$-th robot.
At the inference stage, when each robot needs to calculate the predictive posterior in \eqref{eq:pred}, it fuses all collected models, $M_k$, using a generalised product of experts (GPoE)~\cite{CaoFle2015arxiv},
\begin{equation}
    p_k(f_*|\mathbf{x}_*,t_*,\tilde{\mathcal{D}}^{(k)}) = \prod_{i=1}^{M_k} p_i^{\beta_{i}}(f_*|\mathbf{x}_*,t_*,\mathcal{D}^{(i)}),
\end{equation}
where $\tilde{\mathcal{D}}^{(k)} \subset \mathcal{D}$ are the data sets collected by robot $k$, and each model is weighted by a factor $\beta_{i}$.
We adopt the convention of setting $\beta_{i} = 1 / |M_k|$, which recovers the prior outside the range of the data~\cite{DeiNg2015icml}.
Assuming that all robots will eventually receive the models of all other robots in the swarm, the complete, location-unaware learning procedure is $\mathcal{O}(km^3)$.

\section{EXPERIMENTAL VALIDATION}
We validate LU-GPR using two different scenarios.
First, using a randomly generated synthetic function, we study how the number of robots and the communication range affect estimation quality.
Then, we demonstrate how LU-GPR can be used to model the behaviour of an evacuating crowd.
For simplicity, we follow the common practice of assuming that the means of the GPs are zero in both scenarios~\cite{WilRas20006book}.

\subsection{Simulation environment}\label{sec:sim}
We run all analyses in an environment simulated in Unity.
We consider a cylindrical robotic platform with a 0.25-m radius modelled as a kinematic agent,
\begin{equation}
    \dot{\mathbf{p}}_k = \mathbf{u}_k,
\end{equation}
where $\mathbf{p}_k$ is the position of robot $k$, and $\mathbf{u}_k$ the corresponding control input.
Each robot can detect obstacles ahead and exchange messages with other robots within a distance $r_c$.
We also implement virtual omnidirectional cameras, so that each robot can identify agents in their line of sight, as well as measure their position and velocity, within a range $[r_{v,\min}, r_{v,\max}]$.
For all experiments, each robot executes a ballistic random walk; it moves in a straight line until it encounters another robot or an obstacle, then chooses a new random direction.
This type of random walk has shown good results for exploration tasks with robot swarms~\cite{KegGarBit2019taros}.

Each robot moves at $v_{\max} = 0.5\ \text{ms}^{-1}$, and measures positions and velocities with uncertainties $\sigma_{mp} = 0.02\ \text{m}$ and $\sigma_{mv} = 0.01\ \text{ms}^{-1}$, respectively.
The odometry noise is set to $\sigma_v = 0.1\ \text{m/m}$ (metres per metre travelled).
All other parameters used in the GBP algorithm, including communication and update rates, are set to the same values as those used by Jones and Hauert~\cite{JonHau2025AR}.
For the GPR algorithm, we use the anisotropic squared exponential (ASE) kernel,
\begin{equation}\label{eq:kernel}
    k(\mathbf{x},\mathbf{x}') = \sigma_f^2 e^{-\frac{1}{2}(\mathbf{x}-\mathbf{x}') \Lambda^{-1} (\mathbf{x}-\mathbf{x}')},
\end{equation}
where $\sigma_f = 1.0$ and $\Lambda = \text{diag}([l_1^2,l_2^2])$, with characteristic lengths $l_1=l_2=1.5$.
As we do not consider any model learning in this work, we fix $\sigma_n = 0.1$, $\lambda=0.98$, and the number of random Fourier features to $m=50$.
To obtain the value of $\lambda$, we consider the odometry noise, which increases linearly, as a lower bound for the corresponding, exponentially increasing, GPR forgetting, so that
\begin{equation}
    \lambda \leq \exp\left({- \frac{v_{\max} \sigma_v}{e}}\right),
\end{equation}
where taking the equality results in the value we select.
It is worth pointing out that this is only an approximation, as the odometry noise affects the input variable, $\mathbf{x}$, while the variance of the GP concerns the output variable, $y$.
Each robot samples the target function from the environment every 1 s, and runs the inference update every 5 s.

\subsection{Synthetic function analysis}
\begin{figure}
    \centering
    \includegraphics{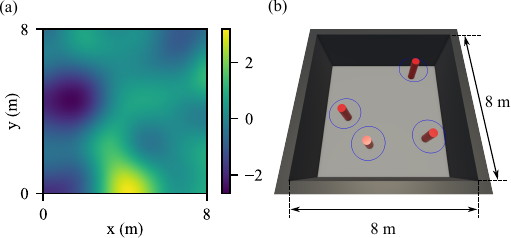}
    \caption{Estimation of a spatial field. (a) Function with zero mean to be estimated. (b) Environmental setup, where a fixed number of robots (here, four) with a limited communication range (here, 1 m, as shown by the blue rings) sample the field as they perform a random walk.}
    \label{fig:toyestimation}
\end{figure}

Fig. \ref{fig:toyestimation} shows the environment constructed to study the effect of the swarm size and the communication radius on the estimation.
We consider a closed, $8\ \text{m} \times 8\ \text{m}$ arena surrounded by walls.
To isolate the behaviour of the LU-GPR algorithm from the model identification problem, we generate a toy function in this arena using a GP with the same parameters as those defined in Section \ref{sec:sim}.
At every inference iteration, each robot computes the posterior in \eqref{eq:pred} on a $20 \times 20$ grid centred in the environment.
In a real-world setting, the robots would not have access to these positions, and would need to select a set of points in their local frame of reference.
However, we use global coordinates to provide a fair assessment of the quality of the estimations.

We consider swarms of 3, 4, 6, and 10 robots, where each robot has a communication radius of 1 m, 2 m, or 4 m.
We also compare the results with a fully connected swarm, where all robots are permanently within communication range of each other.
In this fully connected scenario, the GBP algorithm will converge almost immediately, thus isolating the regression problem from the localisation problem.
For each combination of swarm size and communication range, we execute 10 independent runs, and report the average root mean squared error (RMSE) values corresponding to the estimations of the first robot.
As all robots are initially placed in a random position in the environment, we argue that this selection is arbitrary, and does not affect the results obtained.
Finally, we assume an unweighted approach, where all the samples are given the same, $\rho_i=1$ weight.

\begin{figure*}
    \centering
    \includegraphics{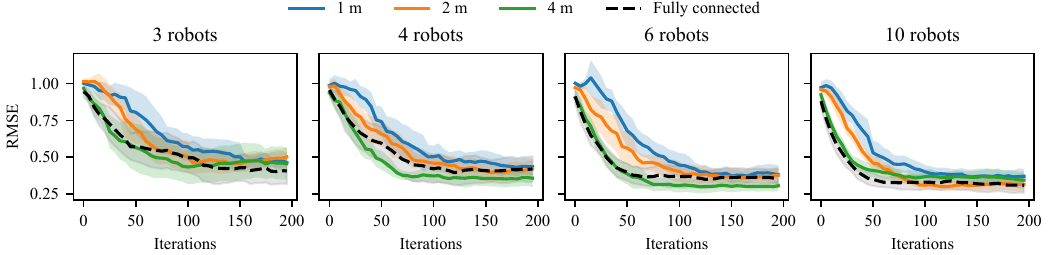}
    \caption{Scalability of LU-GPR. Each plot shows the expected root mean squared error (RMSE) of the global model learned by a single robot per GPR update iteration, obtained by averaging over 10 independent runs. Shaded regions bound the confidence interval with a 0.95 confidence level. We consider local communication with three increasingly large ranges, as well as full connectivity of the swarm.}
    \label{fig:scalability}
\end{figure*}

\begin{figure}
    \centering
    \includegraphics{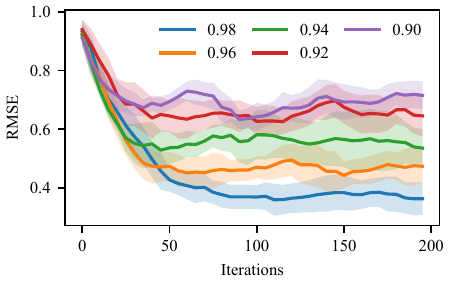}
    \caption{Effect of the magnitude of the forgetting factor, $\lambda$, on the convergence bias. For each value of $\lambda$, we generate 10 independent runs of a four-robot swarm with full connectivity. Each plot shows the expected root mean squared error (RMSE) of the global model learned by a single robot per GPR update iteration, obtained by averaging over the 10 runs. Shaded regions bound the confidence interval with a 0.95 confidence level.}
    \label{fig:forgetting}
\end{figure}

In Fig. \ref{fig:scalability}, we see that the robots converge to a solution, even for small swarm sizes.
Larger swarm sizes lead to reduced estimation bias and variance, and convergence occurs at a faster rate.
Interestingly, the communication range does not appear to have a significant impact on the final estimation; only on the convergence rate.

Despite using the same hyperparameters for function generation and regression, a degree of bias will be inevitable due to the approximation of the kernel function in \eqref{eq:features}.
We also highlight an additional contribution due to odometry; more specifically, the amount of forgetting introduced in the LU-GPR algorithm.
As shown in Fig. \ref{fig:forgetting}, reducing the value of $\lambda$, which increases the amount of forgetting per update, increases the RMSE value to which the robots converge.
Therefore, we infer that larger swarm sizes achieve lower estimation bias through a more persistent monitoring of the environment, where the rate at which samples are forgotten is compensated with the rate at which the system acquires new data.

\subsection{Modelling crowd behaviour}
We now move on to estimating how a crowd of pedestrians moves in an evacuation scenario.
Specifically, our goal is to assess how well a robot swarm can estimate the expected direction of motion of pedestrians throughout the environment.
In an indoor environment, a real robot swarm might not have access to position data.
Therefore, this scenario represents just one case where the swarm needs to exploit the full capabilities of LU-GPR: data correction to obtain consistent posterior estimates while position estimates converge to a common representation, forgetting old data as the accumulated odometry noise increases, and robust estimation to account for uncertain measurements.

\begin{figure}
    \centering
    \includegraphics{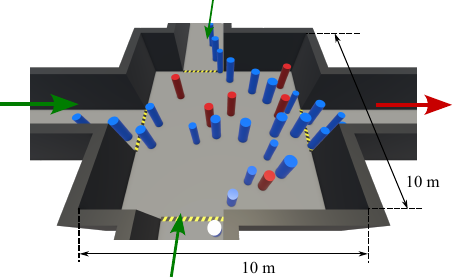}
    \caption{Simulated environment to test crowd monitoring. A swarm of six robots (red cylinders) execute a random walk in a room with pedestrians (blue cylinders) who are entering through three doors (green arrows) and exiting through a single door (red arrow). Black and yellow tape indicates the presence of a doorway.}
    \label{fig:crowdarena}
\end{figure}

To enforce a zero-mean distribution of the data, we construct an environment that is symmetric about the $x$-axis, as shown in Fig. \ref{fig:crowdarena}.
Pedestrians spawn continuously at one of three entrances towards the left-hand side of a $10\ \text{m} \times 10\ \text{m}$ room, and move towards a single exit on the right-hand side.
As a result, we expect all pedestrian orientations to lie in the $[-\pi/2, \pi/2]$ range, and the robots can run a single regression process, instead of the two that would be required for an arbitrary vector field.

We model the behaviour of the crowd using the cognitive model of pedestrian behaviour of Moussa\"{i}d et al.~\cite{MouHelThe2011PNAS}.
We select this model for two main reasons.
First, because it has been shown to accurately replicate both individual and crowd behaviours with various densities, with better results than the commonly used physics-based models~\cite{HelMol1995PRE}.
The second reason concerns the interaction between the swarm and the crowd.
We consider here a passive human-robot interaction scenario, where the robots move freely around the environment, and it is the pedestrians who must avoid colliding with them.
Under these conditions, robots can be regarded as moving agents, and the cognitive model does not need to be modified to account for them.

We set the communication range of the robots to $r_c = 2\ \text{m}$, and their vision range to $[r_{v,\min}, r_{v,\max}] = [0.2\ \text{m}, 2\ \text{m}]$.
To prevent that the robots exit the room during an experimental run, we assume that they can detect the four doorways of the arena with their cameras and avoid them using the same avoidance strategy they use in their random walk.
At every GP-update stage, each robot computes the average direction of motion of the pedestrians they can see.
This is calculated by averaging the normalised velocities of each pedestrian, and taking the orientation of the resulting vector.

To illustrate the benefits of using a robust approach, we compare the performance of the unweighted approach with a weighted one.
For the weighted approach, we set the weight of each sample to the number of pedestrians used to compute the average orientation, as we expect samples obtained from a larger number of pedestrians to be more reliable.

\begin{figure}
    \centering
    \includegraphics{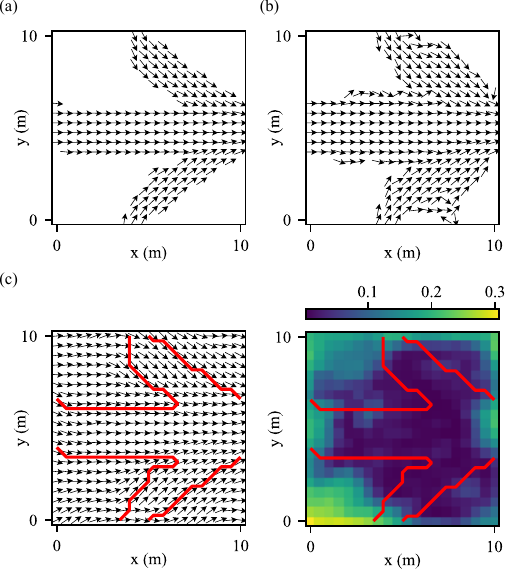}
    \caption{Estimation results of a single robot. (a) Expected pedestrian flow when no swarm is present in the environment. Each arrow is a unit vector representing the mean orientation per cell.
    (b) Expected pedestrian flow when a six-robot swarm is executing a random walk in the environment. (c) Posterior mean (left) and variance (right) obtained at the end of a run by a single robot. 
    The red outline delimits the domain of the expected pedestrian flow without a swarm in the environment.
    }
    \label{fig:crowdrobot}
\end{figure}

To determine how well the estimation compares to the expected pedestrian behaviour, we use as ground truth the average orientation computed in the absence of any robots.
As pedestrians will not pass through all parts of the environment, the domain of the ground-truth data will be smaller than that obtained with the posterior.
Fig. \ref{fig:crowdrobot} illustrates the resulting differences.
While the presence of robots has some effect on the expected pedestrian behaviour around the edges of the domain, the overall patterns remain similar.
Therefore, to compare the posterior means and the ground truth, we rely on the target points that overlap the domain of the ground truth, as shown in Fig. \ref{fig:crowdrobot}c.
We then compute the RMSE by comparing the posterior mean and the ground truth element-wise.
As before, we generate 10 independent runs, and average the results of a single robot.

\begin{figure}
    \centering
    \includegraphics{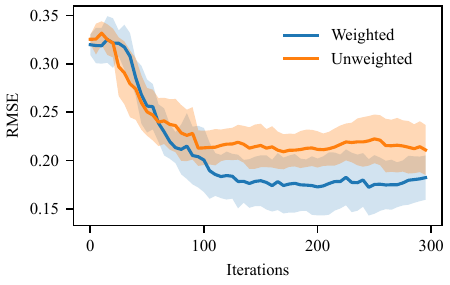}
    \caption{Estimation of crowd orientation by a swarm of six robots, with and without weighting each sample based on the number of pedestrians observed. Each line represents the expected root mean squared error (RMSE) of the global model learned by a single robot per GPR update iteration, obtained by averaging over ten independent runs. Shaded regions bound the confidence interval with a 0.95 confidence level.}
    \label{fig:crowdrmse}
\end{figure}

Fig. \ref{fig:crowdrmse} shows that both the  weighted and unweighted approaches converge to a solution.
The weighted approach appears to improve over the unweighted ones by obtaining lower estimation bias.
However, the high variance resulting from our experimental setting prevents us from making any conclusive statements, as we can see from the overlapping confidence intervals.
For a more visual representation, Fig. \ref{fig:crowdrobot}c illustrates how a single robot can obtain a reliable model of the expected crowd behaviour.

\section{DISCUSSION}
In this work, we have introduced LU-GPR, a solution to the spatial field estimation problem with robot swarms, where the robots do not have access to a positioning system.
To focus on the novel aspects of LU-GPR, we have left several aspects of the GPR framework unexplored, and made some choices that might have led to suboptimal solutions.

First, we have argued that the update rules in \eqref{eq:varupdate} and \eqref{eq:meanupdate} provide and adaptive and robust estimation framework, based on how their mathematical formulations compare to those presented in previous works.
Indeed, our results show that the robots can forget older samples, and that a simple weighting strategy results in a lower estimation bias of crowd behaviour.
However, a systematic assessment of these properties in time-varying scenarios and in the presence of outliers remains an important subject for future work.

There is also the matter of the navigation strategy of the robots.
Here, we have implemented a simple, ballistic random walk.
In order to take advantage of the full power of GPs, it is likely that an exploration strategy that uses the computed posterior variances through acquisition functions could lead to better optimised solutions.
However, conceiving such a strategy is not trivial.
In the absence of full connectivity, robots in a swarm must decide individually how to act.
Should they all use the full posterior estimate to decide where to move next, and assuming that all robots have a sufficiently accurate estimate, all robots could end up choosing the same strategy, resulting in poor exploration.
Therefore, while a random walk might be a suboptimal exploration strategy, it could prove a more robust approach than naive implementations of acquisition functions.

Another aspect is the uncertainty in the input variables, of which we encounter two distinct sources.
The first one concerns the odometry noise.
As we explain in Section \ref{sec:sim}, we use an update rule that results in an exponential increase of noise variance in the output variable.
However, odometry only grows linearly, and in the input variable.
The difference in growth rates leads to a necessarily unrealistic treatment of noise.
Additionally, mapping uncertainties in the input variables to uncertainties in the output variables requires knowledge of the gradient of the latent function.
While, as we show, this does not compromise the ability of a swarm to converge to a solution, a more accurate approach would be to account for uncertainty due to odometry using an errors-in-variables model, which could lead to lower biases.

The second type of input uncertainty comes from the position beliefs obtained through the GBP algorithm.
In LU-GPR, we take the latest position estimates from the factor graph, and ignore the corresponding uncertainties.
As a result, one might argue that the posterior variances are overconfident.
A more accurate treatment of this source of uncertainty is, however, even less trivial than the one coming from odometry estimates.
The reason is that this is a model-wide uncertainty, and mapping it to the individual input data points would destroy the correlation between the samples.
One might think of using it to determine the weights of each individual model in the GPoE model.
However, this would result in a conflicting effect with the uncertainties computed through the forgetting-based updates, and the resulting variances would be too conservative.

Finally, our choice of the GPoE model as an ensembling method could also be questioned.
However, we argue that the GPoE is a suitable approach for decentralised systems, such as robot swarms.
The main reason is that it does not assume any common priors for all the robots in the swarm.
Therefore, each individual robot could, in principle, learn its own hyperparameters.
For example, our approach can directly accommodate existing learning methods for distributed systems, like Bayesian model averaging (BMA).

In short, future work could focus on validating the adaptive and robust nature of the LU-GPR algorithm, enhancing its performance through exploration strategies that exploit the learned models, treat input noise more accurately, and introduce decentralised hyperparameter learning.
In addition, in order to make the solution easily portable to real robots, it would also be worth exploring messaging strategies with lower bandwidth requirements, as well as ensemble learning methods that scale sublinearly with the size of the swarm.
Nevertheless, we have shown that our solution is scalable and robust to both the limited communication ranges of a swarm and varying noise levels in the input data.

\section*{APPENDIX}
To obtain the update rule for $B$ in \eqref{eq:varupdate}, by substituting \eqref{eq:phiupdate} and \eqref{eq:gupdate} into $B = \Phi^\text{T} G^2 \Phi + \sigma_n^2 \Sigma_p^{-1}$, we obtain
\begin{align}
    B_t &=
    \begin{bmatrix}
        R \Phi^\text{T}_{t-1} & \phi_t
    \end{bmatrix}
    \begin{bmatrix}
        \lambda^2 G_{t-1}^2 & \mathbf{0} \\
        \mathbf{0}^\text{T} & \rho_i^2
    \end{bmatrix}
    \begin{bmatrix}
        \Phi_{t-1} R^\text{T} \\
        \phi_t^\text{T}
    \end{bmatrix}
    + \sigma_n^2 \Sigma_p^{-1} \nonumber \\
    &= \lambda^2 R \Phi_{t-1}^T G_{t-1}^2 \Phi_{t-1} R^\text{T} + \rho_i \phi_t \phi_t^\text{T} + \sigma_n^2 \Sigma_p^{-1}.
\end{align}
Adding and subtracting $\lambda^2 R \sigma_n^2 \Sigma_p^{-1} R^\text{T}$, we get
\begin{align}
    B_t &= \lambda^2 R (\Phi_{t-1}^T G_{t-1}^2 \Phi_{t-1} + \sigma_n^2 \Sigma_p^{-1}) R^\text{T} + \rho_i \phi_t \phi_t^\text{T} \nonumber \\
    & \quad +\ \sigma_n^2 \Sigma_p^{-1} - \lambda^2 \sigma_n^2 R \Sigma_p^{-1} R^\text{T} \nonumber \\
    &= \lambda^2 R B_{t-1} R^\text{T} + \rho_i \phi_t \phi_t^\text{T} \nonumber \\
    & \quad +\ \sigma_n^2 \Sigma_p^{-1} - \lambda^2 \sigma_n^2 R \Sigma_p^{-1} R^\text{T}.
\end{align}
As $R$ is an orthogonal matrix, $R^\text{T} = R^{-1}$.
If we assume that the weights are independent, identically distributed, $\Sigma_p = \sigma_p^2 I$, with variance $\sigma_p^2$, then $\lambda^2 \sigma_n^2 R \Sigma_p^{-1} R^\text{T} = \lambda^2 \sigma_n^2 \Sigma_p^{-1}$, and we arrive at \eqref{eq:varupdate}.

For the update rule \eqref{eq:meanupdate}, we proceed in a similar fashion.
Substituting \eqref{eq:phiupdate}, \eqref{eq:gupdate}, and $\mathbf{y}_t = [\mathbf{y}_{t-1}^\text{T}, y_t]^\text{T}$ into $\mathbf{c}_t = \Phi_t^\text{T} G_t^2 \mathbf{y}_t$, we obtain
\begin{align}
    \mathbf{c}_t &= 
    \begin{bmatrix}
        R \Phi^\text{T}_{t-1} & \phi_t
    \end{bmatrix}
    \begin{bmatrix}
        \lambda^2 G_{t-1}^2 & \mathbf{0} \\
        \mathbf{0}^\text{T} & \rho_i^2
    \end{bmatrix}
    \begin{bmatrix}
        \mathbf{y}_{t-1} \\
        y_t
    \end{bmatrix} \nonumber \\
    &= \lambda^2 R \Phi_{t-1}^\text{T} G_{t-1}^2 \mathbf{y}_{t-1} + \rho_i^2 \phi_t y_t \nonumber \\
    &= \lambda^2 R \mathbf{c}_{t-1} + \rho_i^2 \phi_t y_t.
\end{align}

\bibliographystyle{IEEEtran}
\bibliography{bibtex/bib/IEEEabrv,bibtex/bib/misc}

\end{document}